# NEURAL NETWORK APPROACH FOR EYE DETECTION

Vijayalaxmi[1], P.Sudhakara Rao[2] and S Sreehari[3]

[1,2,3]Department of Electronics & Communication Engineering,VITS,Hyderabad
laxmi214@yahoo.co.in
sreehari_ssc@yahoo.com

## *ABSTRACT*

*Driving support systems, such as car navigation systems are becoming common and they support driver in several aspects. Non-intrusive method of detecting Fatigue and drowsiness based on eye-blink count and eye directed instruction controlhelps the driver to prevent from collision caused by drowsy driving. Eye detection and tracking under various conditions such as illumination, background, face alignment and facial expression makes the problem complex.Neural Network based algorithm is proposed in this paper to detect the eyes efficiently. In the proposed algorithm, first the neural Network is trained to reject the non-eye regionbased on images with features of eyes and the images with features of non-eye using Gabor filter and Support Vector Machines to reduce the dimension and classify efficiently. In the algorithm, first the face is segmented using L\*a\*btransform color space, then eyes are detected using HSV and Neural Network approach. The algorithm is tested on nearly 100 images of different persons under different conditions and the results are satisfactory with success rate of 98%.The Neural Network is trained with 50 non-eye images and 50 eye images with different angles using Gabor filter. This paper is a part of research work on "Development of Non-Intrusive system for real-time Monitoring and Prediction of Driver Fatigue and drowsiness" project sponsored by Department of Science & Technology, Govt. of India, New Delhi at Vignan Institute of Technology and Sciences, Vignan Hills, Hyderabad.*

## 1. INTRODUCTION

Recognition of human faces and various features out of still images or image sequences is an actively developing research field. There are many applications with systems coping with problem face and eye detection such as face identification for security purpose. Due to variation in illumination, background, face alignment and facial expression the problem becomes complex.Face and eye recognition is an actively developing research field and one of the most successful applicationsof image analysis and understanding. There is large numberof commercial, securities, gaze and human interaction detection, fatigue detection for intelligent vehicle systems, video conferencing and vision assisted user interface and forensic applications requiringthe use of face recognition technologies. These applicationsinclude face reconstruction, content-based image databasemanagement and multimedia communication. Early eye detection algorithms used simple geometric models, template matching etc.,but the eye detection process has now matured into a science ofsophisticated mathematical representations and matchingprocesses.Major advancements and initiatives in the past ten to fifteenyears have propelled face recognition technology into thespotlight. Eye detection and localization have playedan important role in face recognition over the years. Even greater concern is the structure of image of an eye varies considerably with distance and camera resolution. The motivation of this work is to detect eyes for real time monitoring and prediction of driver fatigue and drowsiness. The following section summarizes various techniques that have been utilized in the field of eye detection research.

## 2. RELATED WORK

A lot of research work has been published in the field of eye detection so far. Various techniques have been proposed using template matching, IR based approaches, feature based approach, Skin detection method, Hough transform method, Eigen space method or combination of these for eye detection.

In the template matching method, segments of input image are compared with previously stored images, to evaluate the similarity using correlation values. The problem with template matching is it cannot deal with eye variations in scale, expression, rotation and illumination. A method of deformable templates is proposed by Yuille et al. This provides an advantage of finding some extra features of an eye like its shape and size at the same time. But the success rate of this method depends on the templates.The most common approach employed to achieve eye detection in real-time [1, 2,3,4] is by using infrared lighting to capture the physiological properties of eyes and an appearance-based model to represent the eye patterns. The appearance-based approach detects eyes based on the intensity distribution of the eyes by exploiting the differences in appearance of eyes from the rest of the face. This method requires a significant number of training data to enumerate all possible appearances of eyes i.e. representing the eyes of different subjects, under different face orientations, and different illumination conditions.Vezhnevets et al. [5] focus on several landmark points (eye corners, iris border points), from which the approximate eyelid contours are estimated. The iris center and radius is detected by looking for a circle separating dark iris and bright sclera. The upper eyelid points are found using on the observation that eye border pixels are significantly darker than surrounding skin and sclera. The detected eye boundary points are filtered to remove outliers and a polynomial curve is fitted to the remaining boundary points. The lower lid is estimated from the known iris and eye corners. In Skin detection method, the detection of the skin region is very important in eye detection.

The skin region helps determining the approximate eye position and eliminates a large number of false eye candidates. There are various color spaces available to detect skin region such as HSV, RGB, YCbCr, and NTSC spaces to increase the efficiency of eye detection. The data was obtained for all the color space components, and was fitted to a Gaussian curve and peak value determines the color of skin. Pentland et al. [7] proposed an Eigen space method for eye and face detection. If the training database is variable with respect to appearance, orientation, and illumination, then this method provides better performance than simple template matching. But the performance of this method is closely related to the training set used and this method also requires normalized sets of training and test images with respect to size and orientation. Another popular eye detection method is obtained by using the Hough transform. This method is based on the shape feature of an iris and is often used for binary valley or edge maps. Hough transform is a general technique for identifying the locations and orientations of certain types of features in a digital image and used to isolate features of a particular shape within an image. Lam and Yuen [6] noted that the Hough transform is robust to noise, and can resist to a certain degree if occlusion and boundary effects. Akihiko Torii and Atsushi Imiya [7] proposed a randomized Hough transform based method for the detection of great circles on a sphere. Cheng Z. and Lin Y [8] proposed a new efficient method to detect ellipses in gray-scale images, called Restricted Randomized Hough transform. The key of this method is restricting the scope of selected points when detecting ellipses by prior image processing from which the information of curves can be obtained. Yip et al. [9] presented a technique aimed at improving the efficiency and reducing the memory size of the accumulator array of circle detection using Hough transform. The drawback of this approach is that the performance depends on threshold values used for binary conversionof the valleys.

Various other methods that have been adopted for eye detection include wavelets, principal component analysis, fuzzy logic, evolutionary computation and hidden markov models. Huang and Wechsler [10] perform the task of eye detection by using optimal wavelet packets for eye representation and radial basis functions for subsequent classification of facial areas into eye and non-eye regions. Filters based on Gabor wavelets to detect eyes in gray level images are used in [11]. Talmi et al. and Pentland et al. [12,13] use principal component analysis to describe and represent the general characteristics of human eyes with only very few dimensions. In[7] Eigeneyes are calculated by applying Karhunen-Loeve-Transformation to represent the major characteristics of human eyes and are stored as reference patterns for the localization of human eyes in video images. Given a new input image of the same size, it is projected into the eigeneye space. The produced vector describes the similarity of this new image to the eigeneyes. If similarity measure (the Euclidean distance between the mean adjusted input image and its projection onto the eigeneye space) smaller than a threshold, the new image is classified as an eye region. Hjelms and Wroldsen [14] utilize Gabor Filters and PCA for eye detection.Li et al. [15] construct a fuzzy template which is based on the piecewise boundary. A judgment of eye or non-eye is made according to the similarity between the input image and eye template. In the template, the eyelid is constructed by a region of adjacent segments along the piecewise boundary. Each segment in the fuzzy template is filled with the darkest intensity value within this segment. This makes the method have invariance to slight change of eye images in size and shape and it gives the method high robustness to different illumination and resolution for input images, by increasing the contrast between eyelid region and its adjacent regions. The proposed algorithm uses a neural network to scan an input window of pixels across the image, where each gray value in the input window serves as an input for the neural network. The neural network is then trained to give a high response when the input window is centered on the eye. After scanning the entireimage, the position with the highest response then reveals the center position of the eye in theimage. In order to allow applicability in far more general situations, i.e. covering a large variety of eye appearances, allowing rotation and scaling, and allowing different lighting conditions the eyes are located by locating micro-features, rather than entire eyes and the neural network responses are post-processed using Gabor filter which exploits the geometrical information about the micro-features.

## 3. PROPOSED ALGORITHM

The proposed algorithm is divided in two stages –training and detection stage. The steps of pre-processing with Gabor filter, and the classification usingNeural Network are executed in both stages. Besides this, in the training the manual segmentation of theregion of interest and the selection of the most significant features through Color models.In the test stage we still have: the automatic extraction of the region of the eyes through the color model and segmentation of the eye candidates by applying morphological operations.

### 3.1 Database

The VITS and GATV face database is used for both stages. It is formed by 10 people with different facial expressions, hair styles, and illumination conditions and without glasses, adding up to 200 grayscale images of different sizes. In the present work, total 100 images are used for training and nearly 100 images for testing.

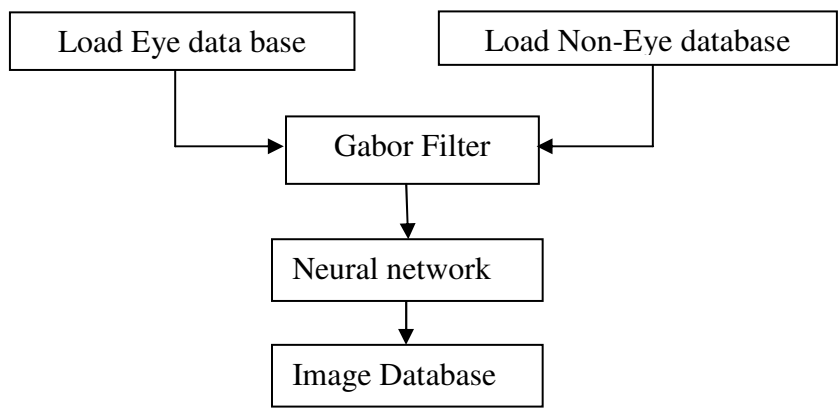

Figure 1: Block Diagram for creating Image Database

### 3.2 Training methodology

To train the neural network in first stage a large number of eye and non-eye images are needed. Nearly 50 eyes cropped from different face examples were gathered from face databases at VITS and from GATV face database. The images contained eyes of various sizes, orientations, positions, open and close. These eye images were used to normalize each eye to the same scale, orientation, and position, as follows:

1. Initialize F F, a vector which will be the average positions of each labeled feature over all the faces, with the feature locations in the first face.
2. The feature coordinates in F F are rotated, translated, and scaled, so that the average locations of the eyes will appear at predetermined locations in a 32x20 pixel window.

Forty eye examples are generated for the training set from each original image, by randomly rotating the images (about their center points) up to $10^0$ scaling between 90% and 110%, translating up to half a pixel, and mirroring.

Practically any image can serve as a non-eye example because the space of non-eye images ismuch larger than the space of eye images. However, collecting a "representative" set of non-eye is difficult. Instead of collecting the images before training is started, the images are collectedduring training, in the following manner, adapted from [13]

1. Create an initial set of non-eye images by generating 50 random images. Apply the preprocessing steps to each of these images.
2. Train a neural network to produce an output of 1 for the eye images, and 0 for the non-eye images.
3. Run the system on an image of scenery *which contains no eyes*. Collect subimages in which the network incorrectly identifies a eye (an output activation > 0).
4. Select up to 40 of these subimages at random, apply the preprocessing steps, and add them into the training set as negative examples. Go to step 2.

Some examples of non-eye that are collected during training are shown in Figure 3. Note thatsome of the examples resemble eyes, although they are not very close to the positive examplesshown in Figure 3. The presence of these examples forces the neural network to learn the preciseboundary between eye and non-eye images. We used 50 images of scenery for collecting negative examples in the manner described above. The eye and non-eye images used for training neural network are show in figure below

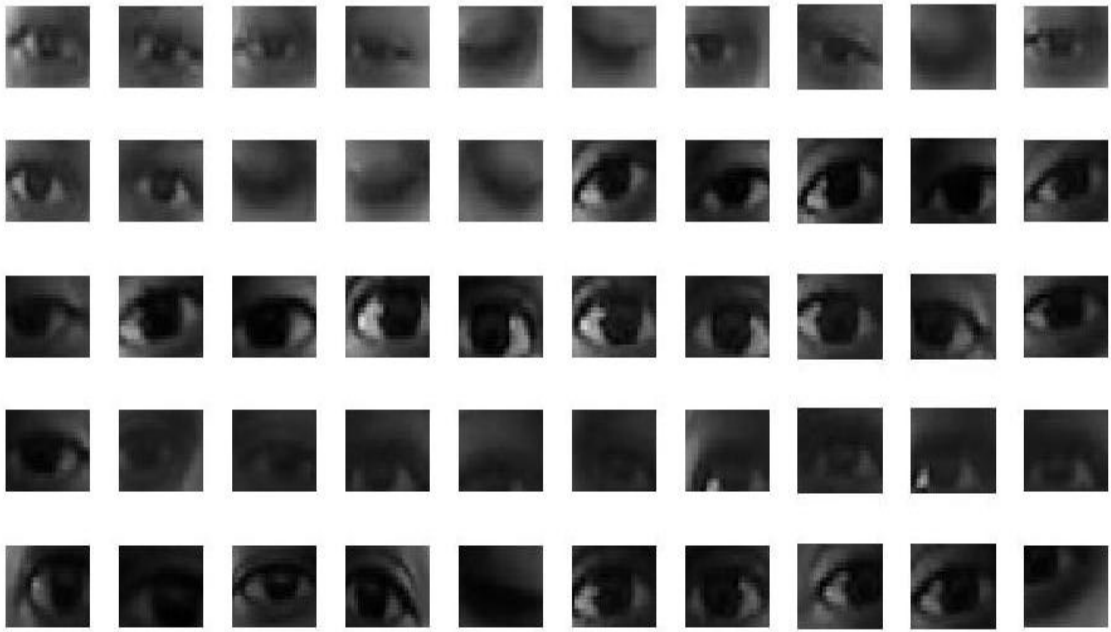

Figure 2: Samples of eye patterns from the training and test sets.

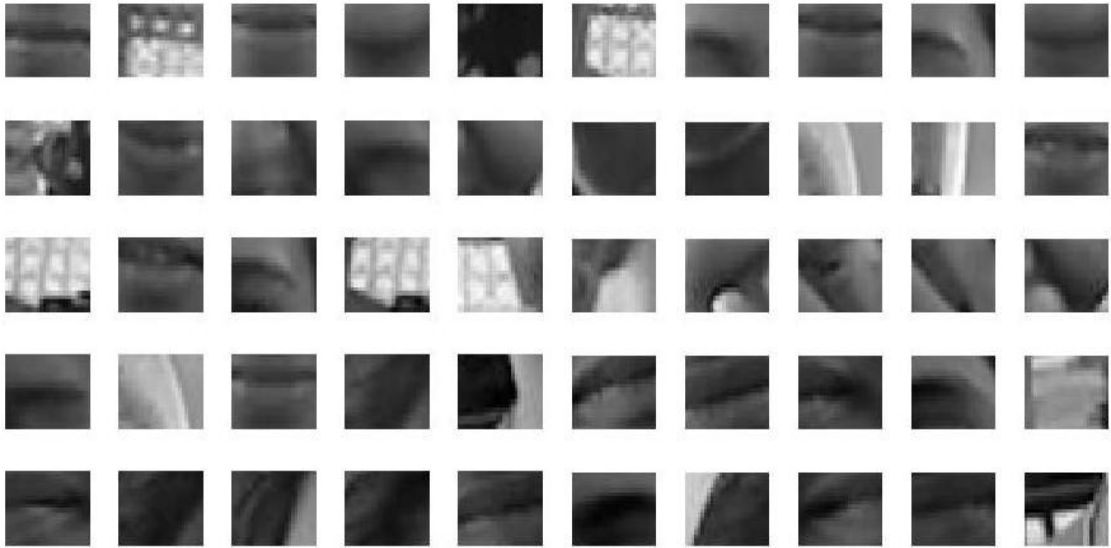

Figure 3: Samples of non-eye patterns from the training and test sets.

## 3.3 Gabor filter

Two-dimensional Gabor functions were proposed by Daugman(1) to model the spatial summation properties (of the receptive fields) of simple cells in the visual cortex. They are widely used in image processing, computer vision, neuroscience and psychophysics. visualize Gabor functions, use a Gabor filter for edge detection and extraction of texture features, simulate simple and complex cells (visual cortex), simulate non-classical receptive field inhibition or surround suppression and use it for object contour detection, and explain certain visual perception effects. The training images are applied to following two-dimensional Gabor function:

$$g_{\lambda,\theta,\sigma,\gamma}(x,y) = \exp\left(-\frac{x'+\gamma^2 y'}{2\sigma^2}\right) \cos\left(2\pi \frac{x'}{\lambda} + \varphi\right) \text{------------------------- (1)}$$

$x' = x \cos\theta + y \sin\theta$
$y' = -x \sin\theta + y \cos\theta$

The Gabor function for the specified values of the parameters "wavelength", "orientation", "phase offset", "aspect ratio", and "bandwidth" will be calculated and displayed as an intensity map image in the output window. (Light and dark gray colors correspond to positive and negative function values, respectively.) The image in the output window has the same size as the input image.

The training data set obtained using Gabor Filter kernel by changing various parameters of Gabor function such as Wavelength, orientation, rotation angle are shown in figure 4 below

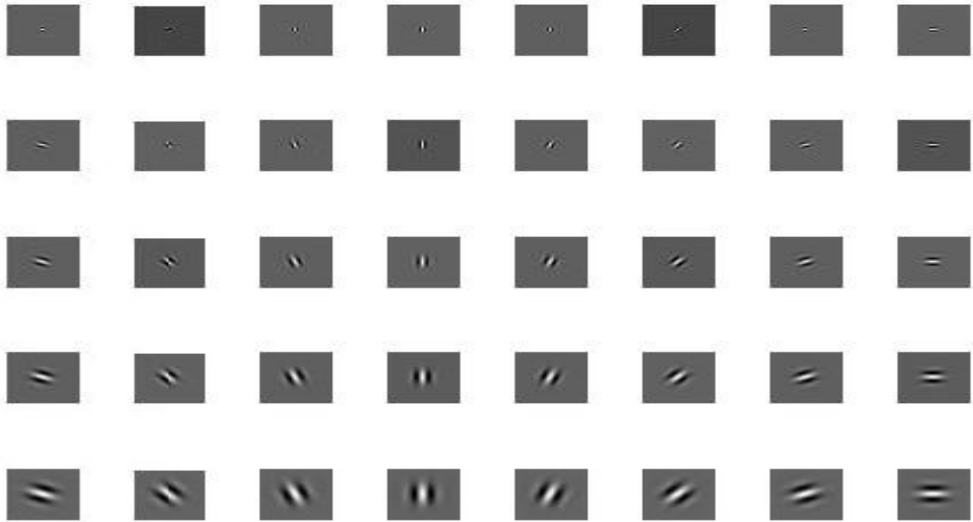

Figure 4: Gabor Filter Kernel with different wavelength, orientation, angle

## 3.4 Neural network

The network will receive the 960 real values as a 960-pixel input image (Image size ~ 32 x 20). It will then be required to identify the eye by responding with a output vector[25]. The output vectors represent a eye or non-eye. To operate correctly the network should respond with a 1 if eye is presented to the network else output vector should be 0 [25]. In addition, the network should be able to handle non-eye. In practice the network will not receive a perfect image of eye which represented by vector as input.

## 3.5 Architecture of neural network

The neural network needs 960 (p1,p2,p3,…..pm) inputs and output layer to identify the eyes. The network is a two-layer log-sigmoid/log-sigmoid network [26], [27]. The log-sigmoid transfer function was picked because its output range (0 to 1) is perfect for learning to output Boolean values [25]. The hidden layer has 200 neurons [25]. This number was picked by guesswork and experience[25]. If the network has trouble learning, then neurons can be added to this layer [25], [28]. The network is trained to output a 1 for correct detection and 0 for non-eye detection. However, non-eye input images may result in the network not creating perfect 1's and 0's. After the network has been trained the output will be passed through the competitive transfer function.

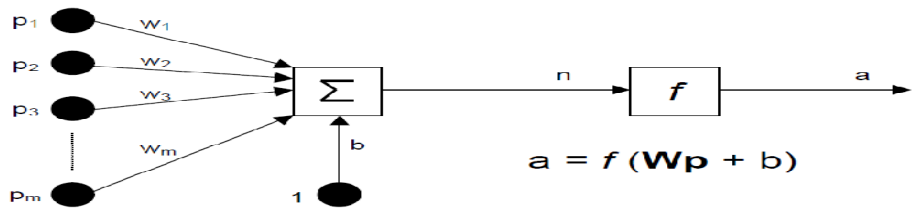

Figure 5: Neural Network Architecture

## 3.6 Test methodology

In the test stage the reliability of the neural network is measured by testing the network with hundreds of input images under different conditions. In the present work an algorithm is developed for segmenting eyes from the given input images. The proposed algorithm is shown in below figure. The algorithm, deals with face extraction using L*a*b Transform color space model by removing the skin region. The eyes are detected by eliminating unwanted regions using HSV Color Space and applying Morphological operation to remove unwanted areas. The proposed algorithm is shown below.

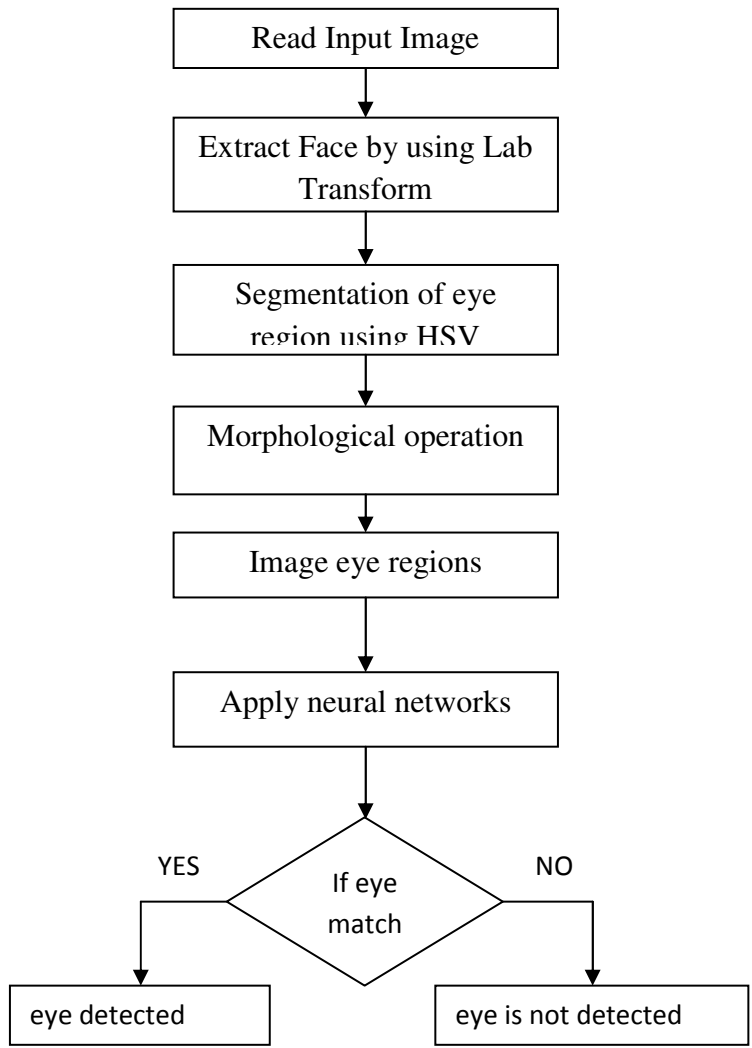

Figure 6: Proposed Algorithm

The first step in this eye detection is segmenting human face from image sequences. In a video simple search techniques will not help to identity the eyes among the other structural components of the face such as beard, moustache,etc. The problem is further complicated with live images, such as video, as the time available to search in each image frame is restricted. Human skin color is an effective feature in face detection. Several different color spaces have been utilized to label pixels as skin including RGB [16], normalized RGB [17], HSV [24], [18], [19], [20], YCbCr [21], YIQ [22], XYZ [23].

The method proposed in this paper involves skin detection to eliminate background components followed by eye detection. Eliminating the skin region helps in determining the approximate eye position by eliminating a large number of false eye candidates. More elaboration was to be made on the skin color algorithms.RGB color space is the most commonly used basis for color descriptions. However each of the coordinates red, green, and blue, does not necessarily provide relevant information about whether a particular image has skin or not as it is subject to luminance effects from the intensity of the image. The Lab color model is used in order to improve color representation. It is a three-dimensional color space in which color differences perceived to be equally large also have equal distances between them. The Lab color model is applied to segment the face. In the proposed algorithm the Lab color model is applied subsequent to applying HSV color space, to get the exact eye region.

The HSV provides color information similar to humans think of colors. "Hue" describes the basic pure color of the image and "saturation" gives the manner by which this pure color (hue) is diluted by white light, and "Value" provides an achromatic notion of the intensity of the color. As proposed in the paper [29] the first two parameters, H and S will provide discriminating information regarding skin. As reported [29] the value of H shall be between 0.01 and 0.1

i.e $0.01 < H < 0.1 \rightarrow$ skin

It may be noted that, this does not remove all the skin in the image but small portions of it remains in the image. As Morphological operations for any image is considered as an advantage. This left out skin areas can be effectively removed using Morphological operations. Finally eyes are segmented from the extracted face using HSV color model and Morphological operations as shown in figure below.The features extracted from the above algorithm are submitted to classification by the already trained SVM. After executing this stage, the results are evaluated using the measurements sensitivity, specificity and accuracy. Sensitivity is defined by TP/(TP + FN), specificity is defined by TN/(TN + FP), and accuracy is defined by (TP + TN)/(TP + TN + FP + FN), where TN is true-negative, FN is false-negative, FP is false-positive, and TP is true-positive.

## 4. EXPERIMENTAL RESULTS AND DISCUSSION

The proposed algorithm is tested onVITS and GATV database. The algorithm is tested on 100 images in the stage of automatic extraction of the region of the eyes. Figure 6 shows successfully detected eye images. On the other hand, in Figure 7 we have examples of images for which the detection failed. We observed that most of the errors occurred due to the position of the face, especially when the face was turning aside, eye closure and distance.The efficiency of proposed algorithm is evaluated on100 images on which the detection of the region of eyes succeeded. The result of the test is with a sensitivity rate of 88.6%, specificity of 95.2% and accuracy of 89.2%. Figure 7shows eye regions which were classified as non-eye. It is noticed that the error occurred in the images whose eyes regions weredarkened, becoming similar to eyebrow, hair or background regions.

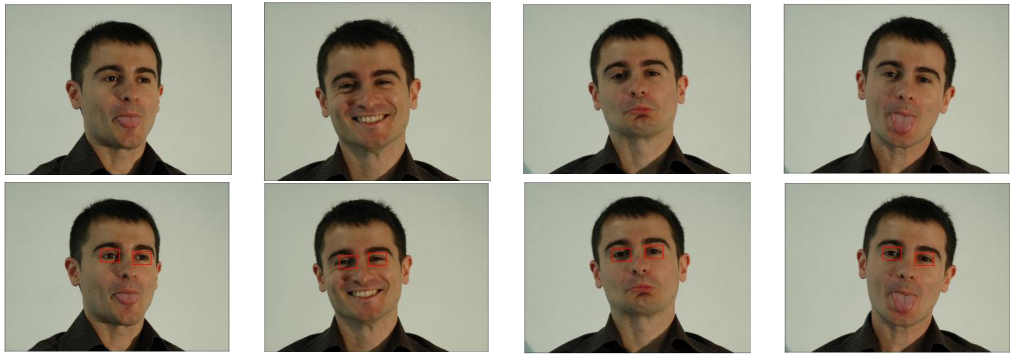

Figure 7: Experimental results showing Original image & Eyes detected

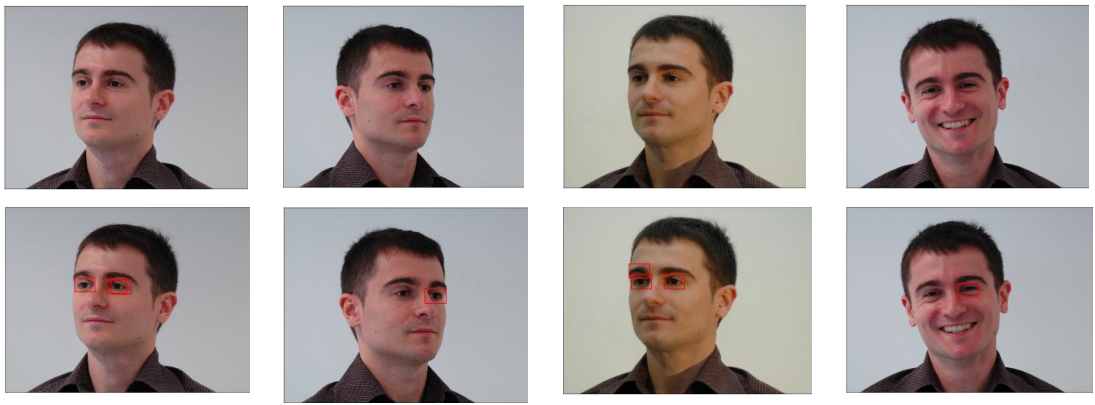

Figure 8: Images with failure in the location of the eyes

## 5. CONCLUSION

Eye detection has become an important issue for different applications. Eyes shall be segmented after eliminating all other portions like background, skin and other human parts in image. Some of the techniques employed by different researches include template matching, Eigenvectors & Hough Transform. We proposed altogether a different technique that employs HSV as well as Lab color spaces for removing all unwanted pixels in the imageexcept eyes and SVM classify the left region in the image as eye or non-eye. Experiments were conducted using 100 images of different head persons, obtained from the standard databases available [24] and VITS database created during the research work. The success rate of the proposed algorithm is 98%, the 2% failure in results is due to head movement, background and due to improper training or neural network.The efficiency of proposed algorithm can also be  evaluatedbased on Sensitivity, Specificity and Accuracy , the calculated values are88.6%,  95.2% and 89.2% respectively. The authors continue to work on the subject for further improvements.

280                    Computer Science & Information Technology (CS & IT)

**Authors**

**First Author Ms.Vijayalaxmi** received B.E Degree from,GNDEC, Visveswaraiah Technological University, Belgaum in 2003. M.Tech. Degree from JNT University, Hyderabad, India in 2008, pursuing Ph.D in the Department of Electronics and Communication Engineering, GITAM University, Vishakapatnam, India. Currently she is working as Associate Professor in the Department of Electronics and communication Engineering, VITS, Hyderabad, India, Her research interest includes Image Processing**.** **Sh**e published 3 papers on "Eye Detection" in international journals and National conferences.

**Second Author Dr.P.Sudhakara Rao** Completed **Ph.D.,**Information and Communication Engineering from Anna University, India, **Masters** in Electronics and Communication Engineering from Anna University, India, **Bachelors** in Electronics and Communication Engineering from Mysore University, India. Worked as deputy **Director, "Central Electronics Engineering Research Institute centre, India"** for over 25 years**, 2 years** as **Vice-president, "Sieger Spintech Equipments Ltd., India",** established an electronic department for the development of electronic systems for nearly 2 years**.** Presently working with Vignan Institute of Technology and Science, Nalgonda District, AP, INDIA as DEAN R&D, HOD ECE. He has one patent and 55 technical publications/ conference papers to his credit. Conducted many international and national conferences as chairman.